\pdfoutput=1

\documentclass[11pt]{article}

\usepackage[preprint]{acl}

\usepackage{times}
\usepackage{latexsym}

\usepackage[T1]{fontenc}

\usepackage[utf8]{inputenc}

\usepackage{microtype}

\usepackage{inconsolata}

\usepackage{graphicx}

\usepackage{amssymb, amsmath}
\usepackage{multirow}
\usepackage{kotex}
\usepackage{CJKutf8}

\title{Bit-level BPE: Below the byte boundary}

\author{Sangwhan Moon \\
  Google LLC \\
  \texttt{sangwhan@iki.fi} \\\And
  Tatsuya Hiraoka \\
  MBZUAI\\
  \texttt{tatsuya.hiraoka@mbzuai.ac.ae} \\\And
  Naoaki Okazaki \\
  Institute of Science Tokyo \\
  \texttt{okazaki@c.titech.ac.jp}\\}

\begin{document}
\maketitle
\begin{abstract}
Byte-level fallbacks for subword tokenization have become a common practice in large language models. In particular, it has been demonstrated to be incredibly effective as a pragmatic solution for preventing OOV, especially in the context of larger models. However, breaking a character down to individual bytes significantly increases the sequence length for long-tail tokens in languages such as Chinese, Japanese, and Korean (CJK) and other character-diverse contexts such as emoji. The increased sequence length results in longer computation during both training and inference. In this work, we propose a simple compression technique that reduces the sequence length losslessly.
\end{abstract}

\section{Introduction}

Byte-pair Encoding (BPE) \cite{sennrich-etal-2016-neural} is a method that allows models to have a robust vocabulary that is capable of representing rare words that have not been seen during training. Variants of this method have been used extensively in many modern natural language processing (NLP) systems, as they allow the representation of a large vocabulary through the concatenation of smaller units, known as subwords, which in turn allows setting an upper boundary on the logits needed for a given model. Byte-level BPE is an extension of BPE to mitigate out-of-vocabulary (OOV). Instead of falling back to OOV when a token cannot be represented through its subwords, it instead encodes the missing token into a sequence of (usually Unicode, in particular - UTF-8) bytes. Due to the introduction of this method, OOV has been largely eliminated in large, foundational models.

However, there is no such thing as a free lunch. Firstly, the model must also learn the intricacies of generating valid Unicode output for any byte-level token, on top of the main linguistic learning (e.g., language modeling) task. For robust generation, training requires enough samples for the model to learn byte sequencing, increasing training costs.  Secondly, representing a character in bytes increases the length by up to four times the input. This limits the utility outside of a large model setup.

\begin{figure}[t]
    \centering
    \includegraphics[width=\linewidth]{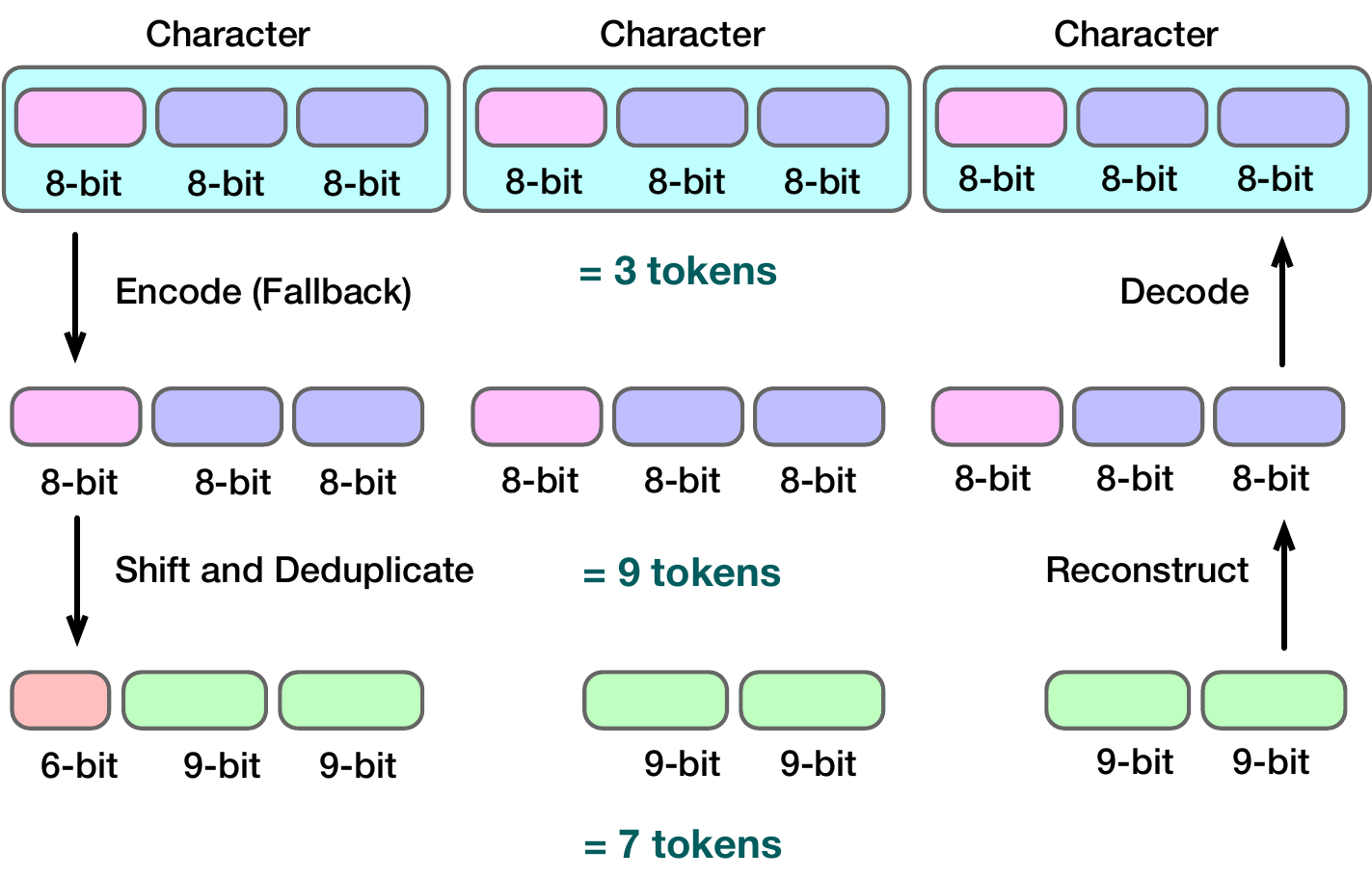}
    \caption{High-level overview of byte-level text representation. In this work, we propose a deduplication method that reduces sequence length at byte level.}
    \label{fig:summary}
\end{figure}

In this work, we investigate the limitations of byte-level fallbacks, particularly the inefficient nature of the byte-level representation. We propose a tokenizer-agnostic method for reducing redundant information in UTF-8 byte-level fallbacks, which results in shorter sequence lengths - eventually saving compute time. Through experiments across three character-diverse languages (Chinese, Japanese, and Korean), we measure the amount of sequence length reduction and assess the pros and cons of our method. To better understand the experiment results and trade-offs, a method to compute the perceived throughput is proposed for comparison.

\section{Background}

\subsection{Byte-level BPE}

Byte-level BPE is an extension of BPE that addresses limitations in dealing with unseen subwords. By adding the lowest-level building block for Unicode characters into the vocabulary, it guarantees that there will be no OOV. While it is unclear where byte-level BPE was initially proposed, mainstream usage began as \citet{Radford2019LanguageMA} and \citet{wang2019neural} proposed it in the context of language modeling and machine translation. The method increases robustness against OOV, which is a common problem in CJK languages.

This problem is due to the sheer diversity of characters needed to represent CJK languages. The CJK Unified Ideographs block defines a total of 97,680 code points, and the Hangul syllables block defines a total of 11,172 code points\footnote{Statistics as of Unicode 15.1.}. Any form of character-level bigram merge, therefore, will result in a combinatorial explosion, which in turn makes supporting any of these languages computationally costly. This complexity, combined with the relatively smaller amount of textual data available in the wild compared to other languages, resulted in many of the CJK characters being treated as OOV. By using byte-level fallback instead of naively losing information (e.g., OOV), modern models have effectively mitigated this problem.

\subsection{Related Work}

Our work is related to the history of character-level and byte-level NLP methods~\cite{zhang2017encoding,el-boukkouri-etal-2020-characterbert,shaham-levy-2021-neural,xue-etal-2022-byt5}.
Although these methods using fine-grained tokenization alleviate the problem of unknown words, they do so at the cost of performance on longer inputs~\cite{libovicky-fraser-2020-towards,gowda-may-2020-finding,goldman2024unpacking}. Byte-level NLP incurs even more severe performance penalties with longer inputs ~\cite{mielke2021between,sreedhar-etal-2023-local}.
\newcite{rust-etal-2021-good} analyzes the effect of over-segmented input, particularly its adversarial effects on model performance.

In the recent LLM era, the fairness of unequal LLM usage costs among different languages is discussed~\cite{ahia-etal-2023-languages,petrov2024language}.
Our proposed method can shorten the input length while preserving byte-level fallback, alleviating disparities in training and inference costs among languages.

\subsection{Byte-level Limitations}
\label{sec:byte-level-limitations}

The clear benefit of using byte-level is that it guarantees there will be no OOV in the final trained model. This benefit, however, comes at the cost of three major caveats.

The first caveat is that one needs to have enough training to ensure that the model does not generate invalid Unicode sequences for a sequence containing rare tokens, as without having seen enough samples, the model will inherently generate invalid Unicode sequences. This results in an implicit requirement to have a larger training dataset to ensure better generalization. This tends to indirectly affect the model size, as the size of the dataset and model should ideally be proportional \cite{hoffmann2022training} - increasing the training cost of the model.

The second is the increased sequence length, which is also inversely proportional to the size of the model's vocabulary\footnote{It is somewhat proportional to the number of languages one needs to support. The more languages supported in a single vocabulary, the more likely there will be a strong dependency on byte-level fallbacks.}. On average, for a CJK character to be represented with UTF-8 bytes, the sequence length grows by three times compared to a character-level representation. The smaller the vocabulary is, the higher the probability of OOV vocabulary is, therefore the sequence length increases. There are tradeoffs to be made here, as a larger vocabulary increases the cost of logit computation and, as a result, requires more computation power for training and inference - but also can introduce undertrained tokens \cite{land2024fishing}.

The last one is a bit more subtle and relates to the tokenizer quality against the byte-level token-only distribution. Tokenizers are often trained with a small sample of the final training corpus, which results in a suboptimal distribution when observed from a whole-corpus perspective. As an unintended side effect, it also creates a local distribution specific to the byte portion, which has distributional characteristics of the underlying character reflect\footnote{In the case of most models, this is UTF-8.}. \citet{zouhar-etal-2023-tokenization} proposes Rényi efficiency as a measure of tokenization quality. This is computed through Rényi entropy $H_\alpha(W_v)$ of the random variable $W_v$ distributed according to $p_v$ for a given vocabulary $V$, defined as: \begin{equation}
H_\alpha(W_v) =\! \lim_{\alpha'\rightarrow \alpha} \frac{1}{1-\alpha'}\log \left( \sum_{w \in V} p_v(w)^{\alpha'} \right)
\label{eq:renyi-entropy}
\end{equation}

and the associated Rényi effiency $E_\alpha(W_v)$ is a scaled with respect to the vocabulary size $|V|$: \begin{equation}
E_\alpha(W_v) \approxeq \frac{H_{\alpha}(W_v)}{\log |V|}
\label{eq:renyi-efficiency}
\end{equation}

In this context, a sharp probability mass distribution of token frequency results in lower Rényi entropy, and intuitively, the more uniform the distribution is, the more entropy increases. Efficiency is a normalized form of entropy, accounting for the vocabulary size $|V|$. In our experiment and tokenizer setup, we can assume that the byte-specific distribution has low entropy, as the frequency disparity between the most frequently-observed bytes and the rest of the vocabulary\footnote{In this case, all 255 possible bytes.} is high, resulting in suboptimal tokenizer efficiency when observed at the byte-level portion of the token distribution. To minimize misunderstandings, we will refer to Rényi efficiency as \textbf{entropy} throughout this paper.

\begin{figure}[t]
    \centering
    \includegraphics[width=\linewidth]{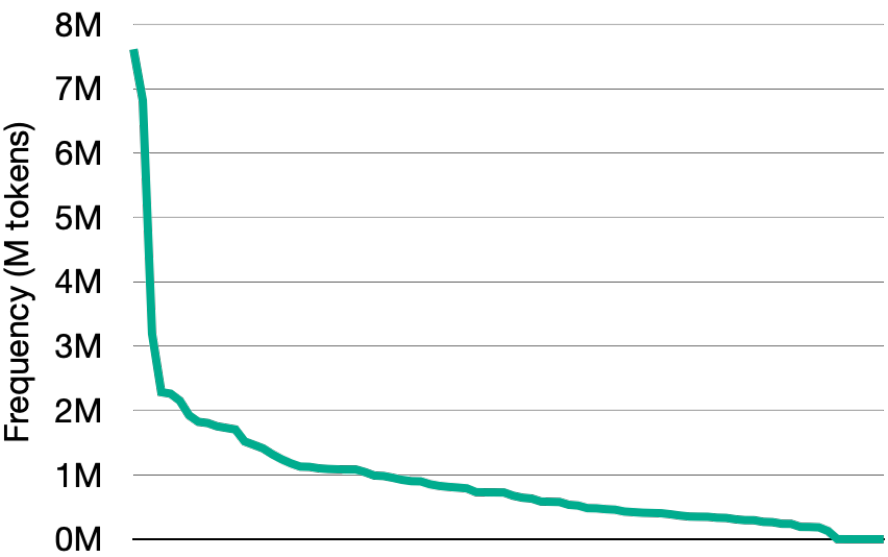}
    \caption{Byte-level frequency distribution on the Japanese-Korean subtitle translation task training set, ranked by most frequent byte.}
    \label{fig:pmf-byte-ranked}
\end{figure}

\section{Proposed Method}
We propose an efficient tokenization method to shorten the input sequence, focusing on the repetition (\S \ref{sec:repetition})  and duplication (\S \ref{sec:duplication}) in UTF-8.
This section first provides the observation nature of the general representation of texts in UTF-8 and then explains the proposed method, which provides the efficient representation (\S \ref{sec:breaking}).
Figure \ref{fig:summary} overviews the proposed method.

\subsection{Duplication in UTF-8}
\label{sec:repetition}

The high-frequency token phenomena observed in Figure \ref{fig:pmf-byte-ranked} is a bias inherited from the underlying character encoding, in this case, UTF-8. For the examples shown in Table \ref{fig:utf8-repetition}, the UTF-8 encoded byte sequence from Chinese, Korean, and Japanese exhibits a duplication problem, indicated in bold. By simply observing this small sample, it is clear that byte-level tokens in the range of \texttt{E4}-\texttt{ED} have a much higher frequency than others.

\begin{table}[t]
    \centering
    \begin{tabular}{|c|c|c|c|}
    \hline
    \multirow{2}{*}{\textbf{zh-CN}} & \begin{CJK*}{UTF8}{gbsn}召\end{CJK*}        & \begin{CJK*}{UTF8}{gbsn}唤\end{CJK*}        & \begin{CJK*}{UTF8}{gbsn}众\end{CJK*}        \\ \cline{2-4} 
                        & \texttt{\textbf{\color{blue} E4} BC 97} & \texttt{\textbf{\color{blue} E5} 94 A4} & \texttt{\textbf{\color{blue} E4} BC 97} \\ \hline
    \multirow{2}{*}{\textbf{ja-JP}} & \begin{CJK*}{UTF8}{min}検\end{CJK*}        & \begin{CJK*}{UTF8}{min}認\end{CJK*}        & \begin{CJK*}{UTF8}{min}裁\end{CJK*}        \\ \cline{2-4} 
                        & \texttt{\textbf{\color{blue} E6} A4 9C} & \texttt{\textbf{\color{purple} E8} AA 8D} & \texttt{\textbf{\color{purple} E8} A3 81} \\ \hline
    \multirow{2}{*}{\textbf{ko-KR}} & 철        & 저        & 히        \\ \cline{2-4} 
                        & \texttt{\textbf{\color{teal} EC} B2 A0} & \texttt{\textbf{\color{teal} EC} A0 80} & \texttt{\textbf{\color{teal} ED} 9E 88} \\ \hline
    \end{tabular}
    \caption{Example text from Chinese, Japanese, and Korean, accompanied with their byte representations. Each character needs three tokens to represent, and we also observe that there are common prefixes.}
    \label{fig:utf8-repetition}
\end{table}

\begin{table}[t]
\centering

    \begin{tabular}{|c|c|c|c|}
    \hline
    \texttt{\textbf{\color{blue} E4}} & \texttt{\textbf{\color{blue} E5}} & \texttt{\textbf{\color{blue} E6}} & \texttt{\textbf{\color{blue} E7}} \\ \hline
    \texttt{\textbf{\color{blue}   111001}00}    & \texttt{\textbf{\color{blue}   111001}01}    & \texttt{\textbf{\color{blue}   111001}10}    & \texttt{\textbf{\color{blue}   111001}11}    \\ \hline \hline
    \texttt{\textbf{\color{purple} E8}} & \texttt{\textbf{\color{purple} E9}} & \texttt{\textbf{\color{purple} EA}} & \texttt{\textbf{\color{purple} EB}} \\ \hline
    \texttt{\textbf{\color{purple} 111010}00}    & \texttt{\textbf{\color{purple} 111010}01}    & \texttt{\textbf{\color{purple} 111010}10}    & \texttt{\textbf{\color{purple} 111010}11}    \\ \hline \hline
    \texttt{\textbf{\color{teal} EC}} & \texttt{\textbf{\color{teal} ED}} & \texttt{\textbf{\color{teal} EE}} & \texttt{\textbf{\color{teal} EF}} \\ \hline
    \texttt{\textbf{\color{teal}   111011}00}    & \texttt{\textbf{\color{teal}   111011}01}    & \texttt{\textbf{\color{teal}   111011}10}    & \texttt{\textbf{\color{teal}   111011}11}    \\ \hline
    \end{tabular}
    
\caption{UTF-8 CJK block common prefixes as seen in bits. E4-E9 are CJK ideographs, while EA-ED is Korean.}
\label{fig:cjk-common}
\end{table}

 As UTF-8 must guarantee robustness to encode every possible character in a deterministic form, duplication is encoded into the scheme so it can be decoded without dependency on the context. This is an important property for reliable transmission and storage, but it has several undesirable properties in the context of text generation from raw bytes. Out of the three undesirable properties discussed in \ref{sec:byte-level-limitations}, the two we intend to address directly in our work are the duplication of information and learning complexity. The length problem is as demonstrated in Table \ref{fig:utf8-repetition}, and the learning complexity can be attributed to the duplication.
 
\subsection{Duplication at Bit-level}
\label{sec:duplication}

Looking at the bit level, the duplication of byte representation can be generalized as the duplication of bit sequences.
We observe that the underlying frequency distribution of the byte-level token range shows overly frequent tokens, supporting our duplication claim. Not only are a third of the bytes redundant information, as shown in Table \ref{fig:utf8-repetition}, but we hypothesize that the redundancy also increases the complexity of learning, as it requires learning the linguistic structure while also learning to generate valid UTF-8 sequences, including emitting the redundant tokens\footnote{A somewhat simple analogy of this is learning how to count while also learning how to play Fizz buzz.}. We also observe a commonality between the most frequent tokens at the bit-level observed in Table \ref{fig:cjk-common}.

Out of the three text samples in Table \ref{fig:cjk-common}, we can observe that aside from the Japanese (ja) sample, the first six bits of the prefix - \texttt{0xEC-0xEF} for Korean (ko), and \texttt{0xE4-0xE7} for Chinese (zh) are shared. This suggests that there is a possibility that this could be considered redundant information that can be de-duplicated.

However, to make this de-duplication possible, we must break a common assumption - that a byte representation, as expressed by the model's tokenizer, is constrained to be eight bits. Having byte-level tokens represented as an exact byte counterpart has an advantage - as long as the model can generate valid byte sequences, it can be decoded without any extra effort. However, this only holds if the model is trained with enough data to learn the linguistic constructs of the corpus and how to generate valid UTF-8. Unless training a large model, this is very challenging, and even in a large model context, it increases the cost of training.

\subsection{Breaking the 8-bit Byte Boundary}
\label{sec:breaking}
\subsubsection*{Overview}
Our method breaks this common assumption, eight bits for a byte,  by treating a sequence of bytes as a sequence of bits. Instead, ignoring the eight-bit boundary and having a flexible boundary of bits allows us to optimize the sequence - in particular, by removing duplicate information. With our method applied, a "byte" in a tokenizer can hold any amount of bits instead of being uniformly eight (see Figure \ref{fig:summary}). This is possible as the model sees the individual bytes as logits, which does not require them to be aligned to eight bits. This optimization would not be possible outside of this context without wasting space.

With bit boundary constraints lifted and the common bits identified, we can now treat the common bits as a shared bit prefix of the upcoming tokens. For example, we can treat \texttt{0b111001} (\texttt{0x39}) to indicate that all following tokens have the same prefix unless the prefix changes or a subword is emitted. The two residual bits (e.g., 01 in the case of E5) will need to be carried over by the following bytes. As the vocabulary allocated for bytes is $2^8$, carrying the two bits over on the next byte will increase the vocabulary budget significantly - as this would mean adding 768 more "byte" representation tokens ($2^{10} - 2^8$). For this reason, we redistribute the bits to have two 9-bit integers for our new extended "byte" representation. This requires adding 256 more tokens. 

We demonstrate this proposed method with the Chinese input shown in Table \ref{fig:utf8-repetition} (\begin{CJK*}{UTF8}{gbsn}召唤众\end{CJK*}):
\begin{center}
    \texttt{\textbf{E4} BC 97 \textbf{E5} 94 A4 \textbf{E4} BC 97} \\
    \label{fig:chinese-example-1}
\end{center}
The process is divided into two parts: encoding and decoding of the text.

\subsubsection*{Encoding}
To isolate the common bits, we choose the first "byte" to be six bits, which results in two residual bits - for example with the character \texttt{E4 BC 97}, the bit boundaries move as follows:

\begin{center}
    \texttt{\textbf{\color{blue}111001}\textbf{00} 10111100 10010111\color{white}\_} \\
    \texttt{\textbf{\color{blue}111001} \textbf{00} 10111100 10010111} \\
    \texttt{\textbf{\color{blue}111001} \textbf{00}1011110 \textbf{0}10010111\color{white}\_} \\
    \label{fig:bit-moving}
\end{center}

After the bit re-distribution, the first six bits are now \texttt{p1=0b111001}, a special token representing a 6-bit prefix\footnote{\texttt{p1=0x39}, which is duplicated across the sequence. In practice, all prefix tokens can usurp existing byte tokens, as they are unreachable. For example, \texttt{0x39} is the character "9".}. This can be implemented through a simple set of bitwise operations; given three bytes $b_1$, $b_2$, and $b_3$ representing a single character, the prefix $\hat{b_1}$, and 9-bit tokens $\hat{b_2}$, and $\hat{b_3}$, are computed by\footnote{$\ll$ and $\gg$ are bitwise left and right shift, respectively.}: \begin{align}
    &\hat{b_1} = (b_1 \land 127) \gg 2 \\
    &\hat{b_2} = (((b_1 \land 3) \ll 7) \lor (((b_2 \land 254) \gg 1) \\
    &\hat{b_3} = ((b_2 \land 1) \ll 8) \lor b_3
\end{align}

By shifting the existing byte boundaries to be six, nine, and nine bits, respectively, we can now represent the same sequence as:

\begin{center}
    \texttt{\textbf{p1} 5E 97 \textbf{p1} CA A4 \textbf{p1} 5E 97} \\
    \label{fig:chinese-example-2}
\end{center}

With naive incremental encoding and a deterministic trailing sequence length for each character\footnote{A prefix is followed by two trailing bytes.}, instances of \texttt{p1} can now be de-duplicated, as the entire sequence is behind the first shared 6-bit prefix. Note that a special token is used instead of the byte representation (0x39) to disambiguate between naturally occurring bytes and incremental decoder triggers. The de-duplicated sequence is:

\begin{center}
    \texttt{\textbf{p1} 5E 97 CA A4 5E 97} \\
    \label{fig:chinese-example-3}
\end{center}

This is a 22.22\% reduction from the original input. Whenever there is a transition in the 6-bit prefix, the new 6-bit prefix \texttt{p$_n$} is emitted. For example, if we append a prefix transitioning Japanese character (\begin{CJK*}{UTF8}{min}認\end{CJK*}) \texttt{E8 AA 8D}, the resulting sequence (\begin{CJK*}{UTF8}{gbsn}召唤众\end{CJK*}\begin{CJK*}{UTF8}{min}認\end{CJK*}) with our method is encoded as:

\begin{center}
    \texttt{\textbf{p1} 5E 97 CA A4 5E 97 \textbf{p2} 55 8D} \\
    \label{fig:chinese-example-4}
\end{center}

With one prefix switch, the gains decrease - 16.66\% shorter than the original input. 

\subsubsection*{Decoding}
As the specific Unicode blocks our method targets have a deterministic length of three bytes per character, we can invert the incremental encoding by emitting the current prefix token after every 9-bit bi-gram, which results in the following generated sequence:

\begin{center}
    \texttt{\textbf{p1} 5E 97 \textbf{p1} CA A4 \textbf{p1} 5E 97 \textbf{p2} 55 8D} \\
    \label{fig:chinese-example-5}
\end{center}

This sequence can be re-aligned to a boundary of eight bits each, which results in a decodable UTF-8 byte sequence, through the following computation: \begin{align}
    &b_1 = (\hat{b_1} \ll 2) \lor (\hat{b_2} \gg 7) \\
    &b_2 = ((\hat{b_2} \land 127) << 1) \lor (\hat{b_3} >> 8) \\
    &b_3 = \hat{b_3} \land 255
\end{align}

This results in the reconstructed sequence:

\begin{center}
    \texttt{\textbf{E4} BB BD \textbf{E5} 81 87 \textbf{E7} AE 80 \textbf{E8} 94 B5} \\
    \label{fig:chinese-example-6}
\end{center}

In our work, the bit boundaries were set to be optimal for CJK scripts - the boundaries can be set differently for different Unicode blocks.

\begin{table}[]
\small
\begin{tabular}{|cl|rrr|rr|}
\hline
\multicolumn{2}{|c|}{\multirow{2}{*}{}}                            & \multicolumn{3}{c|}{\textbf{Length ↓}}                                                                       & \multicolumn{2}{c|}{\textbf{Entropy ↑}}                              \\ \cline{3-7} 
\multicolumn{2}{|c|}{}                                             & \multicolumn{1}{c|}{\textbf{Byte}} & \multicolumn{1}{c|}{\textbf{Ours}} & \multicolumn{1}{c|}{\textbf{Diff}} & \multicolumn{1}{c|}{\textbf{Byte}} & \multicolumn{1}{c|}{\textbf{Ours}} \\ \hline
\multicolumn{1}{|c|}{\multirow{2}{*}{\textbf{en-zh}}} & \textbf{T} & \multicolumn{1}{r|}{131M}          & \multicolumn{1}{r|}{127M}          & 3.13\%                            & \multicolumn{1}{r|}{0.764}         & 0.634                              \\ \cline{2-7} 
\multicolumn{1}{|c|}{}                                & \textbf{B} & \multicolumn{1}{r|}{64M}           & \multicolumn{1}{r|}{60M}           & 6.41\%                            & \multicolumn{1}{r|}{0.586}         & 0.435                              \\ \hline
\multicolumn{1}{|c|}{\multirow{2}{*}{\textbf{en-ja}}} & \textbf{T} & \multicolumn{1}{r|}{176M}          & \multicolumn{1}{r|}{174M}          & 0.83\%                            & \multicolumn{1}{r|}{0.818}         & 0.763                              \\ \cline{2-7} 
\multicolumn{1}{|c|}{}                                & \textbf{B} & \multicolumn{1}{r|}{41M}           & \multicolumn{1}{r|}{40M}           & 3.56\%                            & \multicolumn{1}{r|}{0.580}         & 0.417                              \\ \hline
\multicolumn{1}{|c|}{\multirow{2}{*}{\textbf{ja-ko}}} & \textbf{T} & \multicolumn{1}{r|}{113M}          & \multicolumn{1}{r|}{111M}           & 2.21\%                            & \multicolumn{1}{r|}{0.498}         & 0.485                              \\ \cline{2-7} 
\multicolumn{1}{|c|}{}                                & \textbf{B} & \multicolumn{1}{r|}{59M}           & \multicolumn{1}{r|}{56M}           & 4.25\%                            & \multicolumn{1}{r|}{0.485}         & 0.446                              \\ \hline
\end{tabular}

\caption{Sequence length reductions across the training sets, rounded to the nearest million tokens. T indicates the entire corpus, and B indicates the byte-level portion.}
\label{table:mt-length-efficiency}

\end{table}

\section{Experiments}

To observe the effects of our method, we experimented across three target languages (Chinese, Japanese, and Korean) on a translation task.

\subsection{Datasets}
We used three datasets for this experiment, referenced by their target language for simplicity throughout the paper: English-Chinese (Chinese), English-Japanese (Japanese), and Japanese-Korean (Korean). The Chinese and Japanese datasets used are from WMT20 \cite{barrault-etal-2020-findings}, while the Korean dataset is from AI Hub\footnote{\url{https://aihub.or.kr}}. For the WMT20 shared task, we used a custom split of the WikiMatrix dataset from the news translation task for Chinese (2.3M) and Japanese (3.5M). For Korean, we used a subtitle translation dataset (3M)\footnote{An English-Korean dataset of comparable size was not available for free use as of the time of writing.} Each dataset had a 5K test set held out for evaluation and performance benchmarking and 40K for validation. The remainder was used for training.

\subsection{Model Settings}
Each experiment trains byte-level fallback on a machine translation (MT) model trained from scratch. We use a pre-trained Llama2 tokenizer, which triggers a significant amount of byte-level fallbacks in CJK languages, as can be seen in Table \ref{table:mt-length-efficiency} (see \S \ref{sec:sequence-length-and-entropy}). We compare a Llama2 tokenizer for baseline byte-level BPE (\textbf{Byte}) to an augmented Llama2 tokenizer with our method (\textbf{Ours}). 
We consider this popular tokenizer to be a valid baseline to examine our method, which alleviates the problem in the byte-level representation.
All experiments use identical vocabulary expansion when applying our method - our method requires 256 more byte-level tokens (\texttt{0x100-0x1FF}), along with three prefix tokens (\texttt{p1, p2, p3}). 

The comparison intends to approximate the effects of our method in a from-scratch pre-training setup under a limited compute budget.
Specifically, this experiment aims to demonstrate the challenges of learning byte-level generation on a small model and observe the effects of our method when generating byte sequences compared to a baseline byte-level BPE tokenizer.  Here, we train a 65M parameter vanilla Transformer \cite{NIPS2017_3f5ee243}. 
With the different tokenizations, we trained the model for a fixed number of comparable epochs\footnote{With an exception where we stop the model training if it stalled for 10 epochs in a row.}.

Only translation into the CJK languages is evaluated, as the generation of CJK text has a critical dependency on first being able to generate a valid UTF-8 sequence. We expect the trained model to underperform, as we hypothesize that byte-level learning in a small model context is challenging.

\begin{figure*}[!h]
    \centering
    \includegraphics[width=\linewidth]{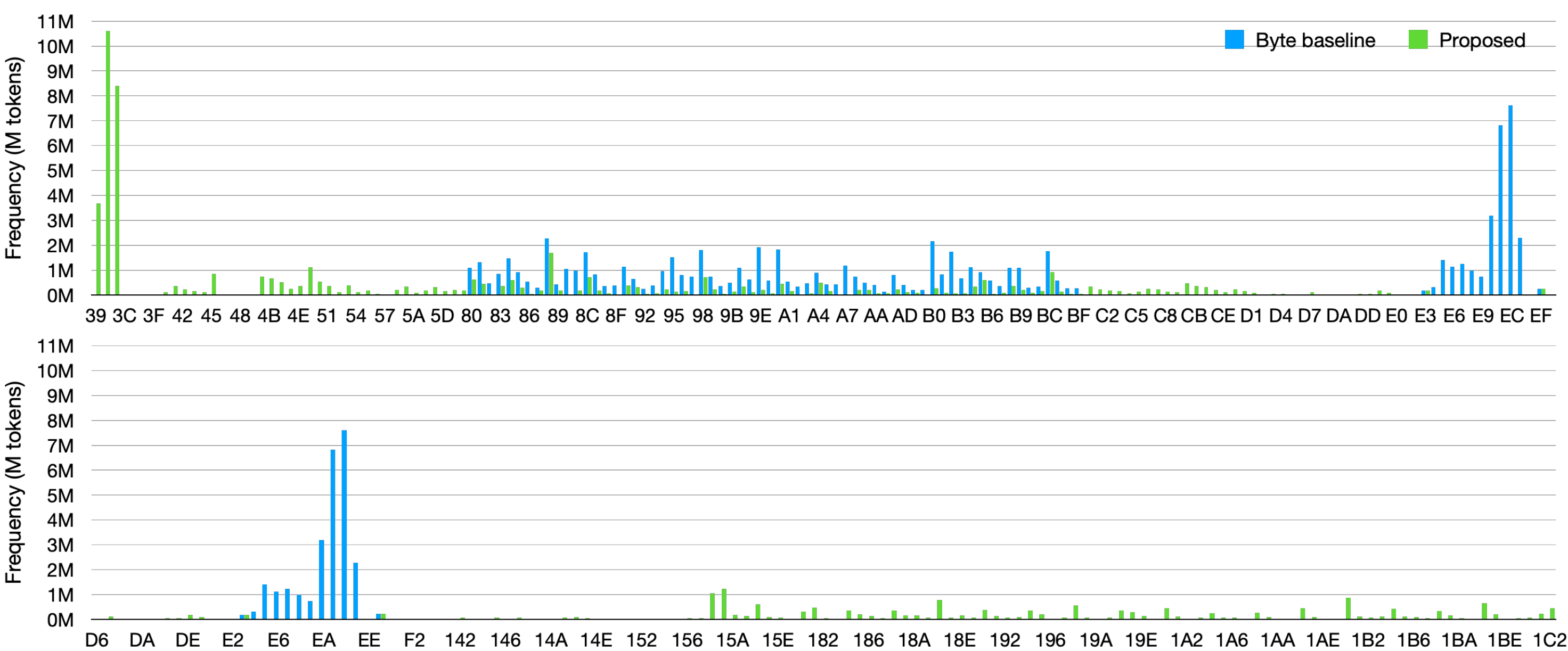}
    \caption{Byte-level distribution shift, with and without our method on the Japanese-Korean subtitle translation task. We can observe that while the probability mass is distributed in our proposed method, there is a sharp increase in the frequency of the most frequent tokens (\texttt{0x39-0x3C}). This results in lower entropy.}
    \label{fig:pmf-bytep-comparison}
\end{figure*}

The model quality was evaluated after reconstructing the byte sequence and performing another detokenization step. With the resulting output, we used sacreBLEU \cite{post-2018-call} under different configurations to compute the translation scores. Different configurations were used for each language, and a unified configuration was used to compute byte-level BLEU and chrF. The sacreBLEU signatures are disclosed in our Appendix. For throughput benchmarks in Table \ref{table:mt-throughput}, we measured the wall time for inference against 5000 samples, with a batch size of 1000\footnote{This is not an entirely realistic setup, as inference time batches will likely not be computed in batches of 1000.}. 

We use a combination of Marian \cite{junczys-dowmunt-etal-2018-marian}, and Huggingface Transformers \cite{wolf-etal-2020-transformers} for these experiments.

\begin{table}[]
\centering
\begin{tabular}{|c|cc|cc|}
\hline
\multirow{2}{*}{} & \multicolumn{2}{c|}{\textbf{Decode error ↓}}       & \multicolumn{2}{c|}{\textbf{Empty ↓}}       \\ \cline{2-5} 
                  & \multicolumn{1}{c|}{\textbf{Byte}} & \textbf{Ours} & \multicolumn{1}{c|}{\textbf{Byte}} & \textbf{Ours} \\ \hline \hline
\textbf{en-zh}    & \multicolumn{1}{c|}{33}           & \textbf{14}           & \multicolumn{1}{c|}{3156}           & \textbf{2545}           \\ \hline
\textbf{en-ja}    & \multicolumn{1}{c|}{136}           & \textbf{0}           & \multicolumn{1}{c|}{218}           & \textbf{87}           \\ \hline
\textbf{ja-ko}    & \multicolumn{1}{c|}{1522}           & \textbf{121}           & \multicolumn{1}{c|}{7}           & \textbf{3}           \\ \hline
\end{tabular}
\caption{Number of invalid output from the 5000 test samples.}
\label{table:mt-problems}
\end{table}

\begin{table*}[!h]
\centering
\begin{tabular}{|r|l|ll|ll|ll|ll|ll|}
\hline
\multicolumn{1}{|l|}{\multirow{2}{*}{}} & \multicolumn{1}{c|}{\multirow{2}{*}{\textbf{Size}}} & \multicolumn{2}{c|}{\textbf{BLEU ↑}}                                    & \multicolumn{2}{c|}{\textbf{chrF ↑}}                                    & \multicolumn{2}{c|}{\textbf{TER ↓}}                                     & \multicolumn{2}{c|}{\textbf{Byte BLEU ↑}}                               & \multicolumn{2}{l|}{\textbf{Byte chrF ↑}}          \\ \cline{3-12} 
\multicolumn{1}{|l|}{}                  & \multicolumn{1}{c|}{}                               & \multicolumn{1}{c|}{\textbf{Byte}} & \multicolumn{1}{c|}{\textbf{Ours}} & \multicolumn{1}{c|}{\textbf{Byte}} & \multicolumn{1}{c|}{\textbf{Ours}} & \multicolumn{1}{c|}{\textbf{Byte}} & \multicolumn{1}{c|}{\textbf{Ours}} & \multicolumn{1}{c|}{\textbf{Byte}} & \multicolumn{1}{c|}{\textbf{Ours}} & \multicolumn{1}{l|}{\textbf{Byte}} & \textbf{Ours} \\ \hline \hline
\textbf{en-zh}                          & 2.23M                                               & \multicolumn{1}{l|}{0.5}           & \textbf{1.7}                       & \multicolumn{1}{l|}{1.9}           & \textbf{3.2}                       & \multicolumn{1}{l|}{127.3}         & \textbf{100.4}                     & \multicolumn{1}{l|}{3}             & \textbf{7.3}                       & \multicolumn{1}{l|}{3}             & \textbf{5.5}  \\ \hline
\textbf{en-ja}                          & 3.47M                                               & \multicolumn{1}{l|}{1.6}           & \textbf{3.2}                       & \multicolumn{1}{l|}{6.5}           & \textbf{8.5}                       & \multicolumn{1}{l|}{467.2}         & \textbf{100.3}                     & \multicolumn{1}{l|}{15.2}          & \textbf{20}                        & \multicolumn{1}{l|}{19.4}          & \textbf{23.5} \\ \hline
\textbf{ja-ko}                          & 2.99M                                               & \multicolumn{1}{l|}{19}            & \textbf{24.6}                      & \multicolumn{1}{l|}{\textbf{35.8}} & 35.5                               & \multicolumn{1}{l|}{114.4}         & \textbf{100.2}                     & \multicolumn{1}{l|}{49}            & \textbf{53.3}                      & \multicolumn{1}{l|}{\textbf{54.7}} & 53.5          \\ \hline
\end{tabular}
\caption{Translation performance results across the three language pairs from 5000 test samples. Byte BLEU and Byte chrF operates were computed at the byte level after reconstructing the output sequence. The table heading ↑ indicates metrics where higher is better, while ↓ it indicates metrics where lower is better.}
\label{table:mt-results}
\end{table*}

\subsection{Results}

\begin{table*}
\centering
\begin{tabular}{|r|cc|cc|cc|}
\hline
\multicolumn{1}{|c|}{\multirow{2}{*}{\textbf{}}} & \multicolumn{2}{c|}{\textbf{en-zh}}                  & \multicolumn{2}{c|}{\textbf{en-ja}}                  & \multicolumn{2}{c|}{\textbf{ja-ko}}                  \\ \cline{2-7} 
\multicolumn{1}{|c|}{}          & \multicolumn{1}{c|}{\textbf{Byte}} & \textbf{Ours}   & \multicolumn{1}{c|}{\textbf{Byte}} & \textbf{Ours}   & \multicolumn{1}{c|}{\textbf{Byte}}   & \textbf{Ours} \\ \hline \hline
\textbf{Tokens Out}             & \multicolumn{1}{c|}{188,361}        & 194,891          & \multicolumn{1}{c|}{578,446}        & 465,401          & \multicolumn{1}{c|}{168,233}          &   302,867      \\ \hline
\textbf{AvgTok Out}             & \multicolumn{1}{c|}{33.64}         &     60.57       & \multicolumn{1}{c|}{115.69}        & 93.08           & \multicolumn{1}{c|}{37.67}           & 38.98         \\ \hline
\textbf{Total time (s)}             & \multicolumn{1}{c|}{72.41}         & 201.65           & \multicolumn{1}{c|}{529.05}        & 271.60           & \multicolumn{1}{c|}{41.10}           & 41.59         \\ \hline
\textbf{Tokens per Second (TPS)}          & \multicolumn{1}{c|}{464.69}        &     300.39      & \multicolumn{1}{c|}{183.91}        & 342.71          & \multicolumn{1}{c|}{916.43}          & 937.21        \\ \hline
\textbf{Tokens in test reference}            & \multicolumn{1}{c|}{291,857}             & 282,423          & \multicolumn{1}{c|}{253,540}             & 251,261         & \multicolumn{1}{c|}{189,800}               & 185,628        \\ \hline
\textbf{Relative Gain}            & \multicolumn{1}{c|}{1}             &     1.0334      & \multicolumn{1}{c|}{1}             & 1.00091         & \multicolumn{1}{c|}{1}               & 1.0225        \\ \hline
\textbf{Perceived TPS}        & \multicolumn{1}{c|}{\textbf{464.70}}        & 310.43 & \multicolumn{1}{c|}{183.91}        & \textbf{345.81} & \multicolumn{1}{c|}{916.43} &     \textbf{958.28}    \\ \hline
\end{tabular}

\caption{Perceived TPS of a model. Tokens out are the total tokens output during the test, and AvgTok is the total divided by the test set size (5K). Tokens per Second were computed with the total tokens divided by the mean runtime across 5 runs. Relative gain is the perceived TPS improvement by reduced sequence length.}
\label{table:mt-throughput}
\end{table*}

\subsubsection*{Sequence length and Entropy}
\label{sec:sequence-length-and-entropy}
Table \ref{table:mt-length-efficiency} reports the sequence length of the texts tokenized with the original Llama2 tokenizer (Byte) and the one with the proposed method (Ours).
This table demonstrates that the Llama2 tokenizer yields a lot of byte-level tokens for each dataset (e.g., 64M out of 131M tokens are byte-level in the ``en-zh'' dataset tokenized with the baseline).
This result also shows that the proposed method shortens the length of sequence across all CJK datasets with the de-duplication technique, which results in increased computation efficiency.
The magnitude of sequence reduction was higher in Chinese and Korean. The reason for this can be explained through the proportion of byte fallback, Japanese only has 23.4\% byte fallback, while Chinese has 48.8\% and Korean as 52\%. This sequence length comes at a cost - the immediate side effect being an increase in parameter count. 

Table \ref{table:mt-length-efficiency} shows changes in entropy as computed by Renyi efficiency, defined in equation \eqref{eq:renyi-efficiency}. As our method's application decreases tokenization's entropy, we can assume that there is a model quality degradation - as \citet{zouhar-etal-2023-tokenization} found entropy correlates to quality. In particular, as observed in Table \ref{table:mt-length-efficiency} and Figure \ref{fig:pmf-bytep-comparison}, a trade-off between entropy and sequence length needs to be made. In our work, we prioritize sequence length over entropy.

\subsubsection*{Reduction of Decoding Error}
The number of sequences that were unable to be decoded see a significant decrease, as can be observed in Table \ref{table:mt-problems}. We also see a significant decrease in empty sequences, although, for Chinese, more than half of the generated output was empty\footnote{The initial model (200 epochs) was much worse, with 95\% of the output of the 5K test samples given translated to blank text on both byte and ours.}. This required us to train the Chinese model for 100 more epochs, significantly reducing the amount of blank output. Other languages tended to have many invalid sequences; Korean had a higher tendency to produce invalid byte sequences over blank sequences, while Japanese suffered the least. This is likely because Japanese has the best in-vocabulary coverage out of the three languages, requiring less dependency on byte-level fallbacks, as seen in Table \ref{table:mt-length-efficiency}. These results suggest that byte-level generation is challenging for smaller model setups.

\subsubsection*{Task Performance}
Table \ref{table:mt-results} shows the translation performance results. The first thing that is very clear is that the BLEU scores are quite distant from public baselines based on the trained models. In particular, Chinese and Japanese are unlikely to be meaningful candidates for quantitative comparison. For this reason, we will focus on the Korean model, where we can perform a better analysis. For the Korean task, we see modest gains between our proposed method and the byte baseline in every aspect. However, this comes at the cost of longer computing times. We see gains for Chinese and Japanese, but both models are likely to be severely undertrained.

We hypothesize that using the Transformer architecture with the default sequence length, long inputs, and byte-level tokenization may have contributed to the disappointing BLEU scores. The primary difference between the cases where we observe positive signals compared to the undertrained model cases is the average length per sample - as subtitles tend to have much shorter sentences.

\section{Factoring Tokenization into TPS}

Table \ref{table:mt-throughput} reports the statistics about the inference time in each dataset.
Initially, we naively assumed that sequence length is linearly correlated to the wall clock time needed by the accelerator to compute the sequence. This did not turn out to be true, and the extra token cost especially needs to be factored in, as it introduces extra computation with an increase in parameters. There are also extra costs when computing the logits and attention. This means one needs to consider the gains that come from reducing the sequence length with our method and whether the extra compute cost compared to the gains results in a net loss. This was one area where there was not much existing literature, which required us to invent an approximation method of the perceived TPS.

Perceived TPS, simply put, is the time it takes to convey the same amount of text to a recipient with different tokenization and model configurations. The coefficients needed to approximate the perceived TPS are in Table \ref{table:mt-throughput}. The most common measure of throughput used is tokens per second (TPS), which is a reliable measure of the model's output speed. However, it does not consider the inefficiencies stemming from poor tokenization. A byte-heavy output can require around twice the amount of tokens generated to transmit the same amount of textual information compared to that of a model that does not use bytes. Our method reduces the sequence length, reducing the amount of tokens needed to convey the same information.

To compute perceived TPS, we introduce the concept of \textit{relative gain}, a simple TPS multiplier that factors in the expected differences between the two tokenizers of models with identical TPS. This is computed as the relative sequence length of the same text between two tokenizers. The assumption made here is that given the same text, a tokenizer that is capable of encoding it with less tokens has higher expressive power, therefore when used in conjunction with TPS results in a better throughput score of the model. It is computed by $|T_c| \over |T_e|$, where $T_c$ and $T_e$ are tokenized sequences using the control and experimental tokenizers, respectively.

For our results, we used the test set for each language to compute the gain for our case, assuming optimal output. This was a conscious decision, as the relative gain computed by a poorly performing model output tended to overamplify the gains\footnote{This was mostly caused by repeated output in undertrained model output.}.

For example, the gain of our method on the Korean translation task is 1.0334, which suggests that there is a 3.34\% advantage in expressive power with the same token count compared to the baseline. This can be multiplied by TPS better to approximate the model and tokenizer combination's perceived TPS. In the last row of Table \ref{table:mt-throughput}, perceived TPS can be compared to conventional TPS.

\section{Conclusion}

Byte-level representation is a simple yet effective method for ensuring full vocabulary coverage for arbitrary Unicode input. However, it comes at the cost of longer sequences, which increase computation time in training and inference. Additionally, it has a high risk of generating invalid sequences, especially with smaller models and training datasets. Our work shows that the effects of these undesirable properties can be partially mitigated through a minor re-encoding of input and output data. Our proposed method is validated by demonstrating its efficacy in reducing sequence length and the number of failed decode operations. The positive results demonstrate that raw UTF-8 bytes are a suboptimal representation and suggest further investigation into an alternative method for text representation, especially at the sub-byte scale.

While evaluating this work, we discovered that TPS, as a measurement of model throughput, is not representative of throughput when seen from a completed text output perspective. We hope that in future model throughput discussions, factoring in tokenization through relative sequence length between models will be used in conjunction with TPS. As it is presented today, the findings in this work have a theoretical trade-off concerning tokenizer entropy and sequence length. Additionally, the work is still constrained to the underlying UTF-8 framework. This is still far from optimally handling long-tail tokens. We expect our findings to be useful as an incremental step toward further challenging the status quo of naively using UTF-8 for long-tail token coverage.

\section{Future Work}
\label{sec:todo}

Byte-level and particularly alternative encoding methods are underexplored as of today, and many promising avenues of investigation remain.

In this work, the bit spans are fixed for implementation simplicity reasons. This is an acceptable first step to validate this method for the context of our experimental setup, scoped to CJK languages. However, this can be further expanded as an optimization problem - minimizing sequence length, minimizing the amount of extra tokens needed, and maximizing entropy. Our work does not address the problem of finding optimal bit boundaries.

Our method decreases tokenization entropy, likely degrading performance. This might be fixable by decreasing the frequency of the most frequent tokens. For example, one could omit the prefix if the previous subword shares the same prefix.

Additionally, the increase in model size caused by the extra vocabulary can be further optimized by utilizing unreachable and, therefore, untrained byte-level tokens. We considered this a premature optimization and did not explore this direction.

Finally, the utility of perceived TPS was only investigated for the scope of our work. We expect this to be more useful when applied to configurations with a larger delta, such as when comparing Llama2 (32K) and Llama3 (128K), for example.


\section*{Limitations}

Our work is an early investigation in a previously underexplored area of tokenization. As this is the first step in a new direction for byte-level tokenization methods, we focused mostly on the paradigm shift of disregarding byte boundaries. As noted in Section \ref{sec:todo}, the investigation was simplified by restricting the scope to CJK languages, allowing us to have a fixed boundary. This particular split may not be optimal for other scripts.

Dataset limitations and compute costs were huge factors that limited the scope of our investigation. CJK languages are comparatively higher-resource, compared to scripts in some other larger blocks - such as Tangut (6,904 characters) or Yi (1,220 characters) where resources are scarce in terms of both datasets and raw web corpora. Emojis are another area we excluded from the investigation, as they are likely not as frequent as text and, therefore, unlikely to benefit as much from deduplication.

\section*{Ethical Statement}

Our method itself does not introduce any new ethical risks when used in conjunction with existing methods. The method is data and task-agnostic and is a method of lossless sequence compression. There are likely minor environmental benefits that come from the reduced computing costs.

The models from our experiments carry the same ethical risks and bias as the underlying dataset used to train them. To minimize the risk, we resorted to well-known or openly available datasets whenever possible and, therefore, can state that we are introducing no novel vectors of societal harm through this work.

\bibliography{anthology,custom}

\appendix

\section{Appendix}
\label{sec:appendix}

\subsection{Fine-tuning and Undertrained Tokens}

As a negative result, we also ran experiments to observe the effects of our method when applied to fine-tuning or continual pre-training of a model. In this experiment, we fine-tuned a Llama2 7B model to observe the effects of our method\footnote{This experiment had to be abandoned due to compute budget constraints.}. This experiment intends to approximate the effects in a foundational model setting by fine-tuning a model through low-rank adaptation \cite{hu2022lora}. We expected this to have characteristics similar to those of pre-training a model from scratch.

Here, we apply our method to the tokenization stage of a pre-trained Llama2 7B model. For newly added tokens, we apply an embedding copy method. This is for better initialization compared to a randomly initialized embedding. A Japanese instruction-tuning dataset with 9.07M training instances \cite{Hirano2023-llmj} was used to fine-tune the pre-trained model.

The embeddings are copied from the existing byte-level tokens with the smallest hamming distance to the newly added token. For example, \texttt{0x1AF} will be initialized with the same embedding as \texttt{0xAF}. The prefix tokens \texttt{p1, p2, p3} get a slightly different treatment, where the embedding is an average across the four prefixes the new prefix token represents. For example, \texttt{p1} will be initialized by the mean between the embeddings for \texttt{0xE4-0xE7}.

\begin{figure*}[!h]
    \centering
    \includegraphics[scale=0.3]{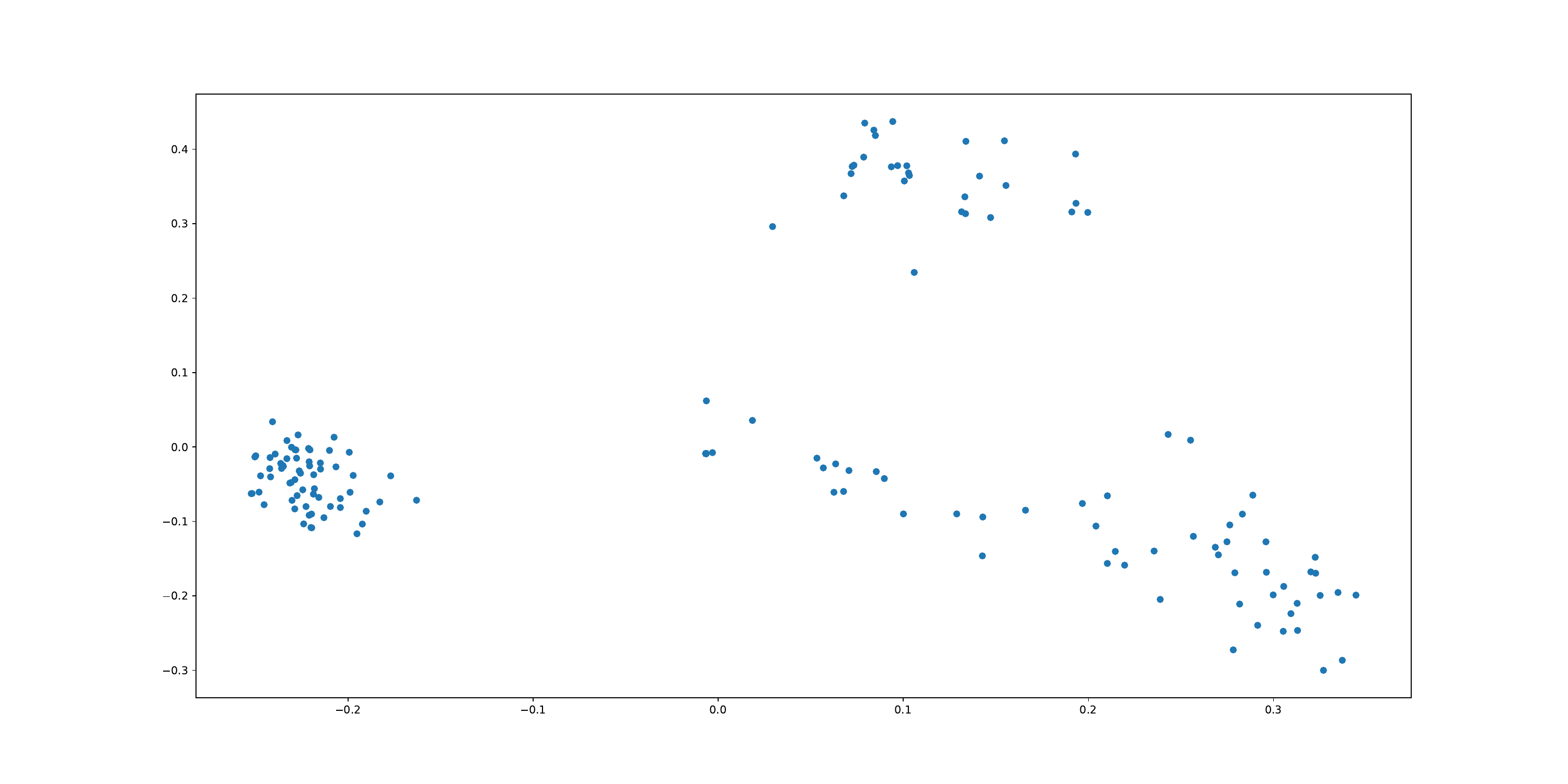}
    \includegraphics[scale=0.3]{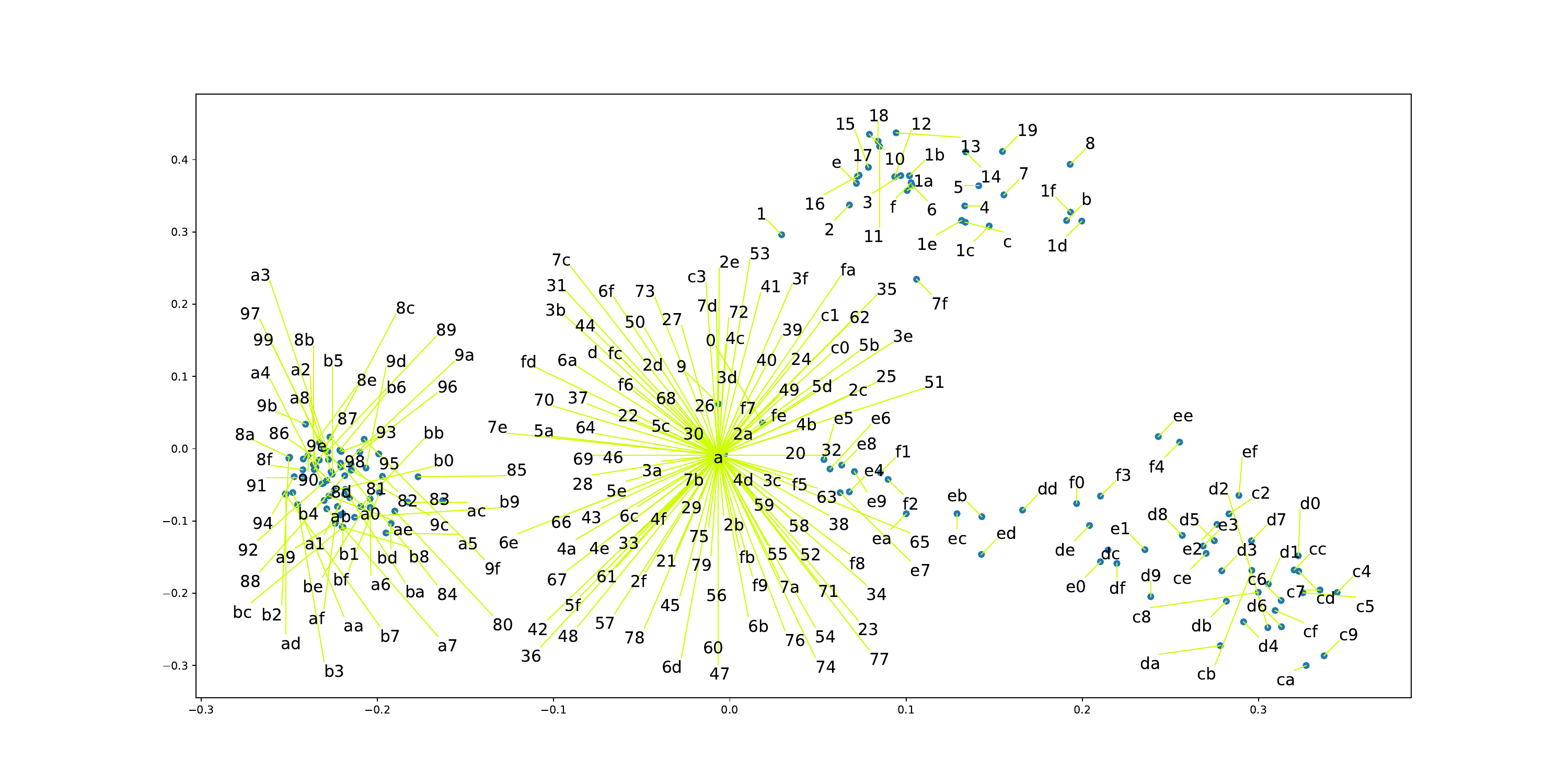}
    \caption{Byte embeddings in Llama2, dimensionality reduced with PCA ($d=2$). We observe there are a couple clusters formed here, along with a dense cluster of undertrained tokens. The undertrained tokens in the projected space converge around the unit vector, and the number of underused tokens can be observed by the number of labels pointing to the unit vector in the lower diagram.}
    \label{fig:byte-embeddings}
\end{figure*}

This model did not produce any meaningful results after training for 20K steps, which can potentially be attributed to severely undertrained tokens being copied over, causing large losses and, therefore, catastrophic forgetting in the initial steps. We later confirmed this through inspection, and the undertrained tokens can be observed in Figure \ref{fig:byte-embeddings}.

However, this suggests that these unreachable tokens can be recycled for the newer 9-bit "bytes" (\texttt{0x100-0x1FF}) if the parameter budget needs to be optimized.

\subsection{Environment and Training Setup}

All of the translation experiments were performed on a shared environment, using two Nvidia H100 HBM2 (94GB). The LoRA experiment was done in the same environment, using four of the same compute accelerators. At inference time, all experiments involving wall-clock measurements were done using a dedicated compute node with a single Nvidia A6000 (48GB).

Korean translation training was run for 24 hours, Japanese for 72 hours, and Chinese for 96 hours (H100x2). The LoRA experiment (H100x4) was run for 120 hours. Performance benchmark inference runs (A6000x1) were run over the course of 60 hours.

Each model was trained initially for 200 epochs\footnote{The early stop criteria until training stalled for 10 epochs in a row.}. However, the Chinese model output was mostly (95\%+) empty at 200 epochs and was trained for another 100 epochs.

\begin{itemize}
    \setlength\itemsep{-0.4em}
    \item \textbf{en-zh} byte: 296 epochs
    \item \textbf{en-zh} ours: 296 epochs
    \item \textbf{en-ja} byte: 200 epochs
    \item \textbf{en-ja} ours: 201 epochs
    \item \textbf{ja-ko} byte: 93 epochs
    \item \textbf{ja-ko} ours: 93 epochs
\end{itemize}

\subsection{Artifacts and Licensing}

In the scope of this work, we created a reference implementation and a pre-trained model as scientific artifacts. The reference implementation and pre-trained models will be distributed under the MIT license at [To be populated after CR]. The reference implementation will not contain the Llama2 tokenizer with our method, but an implementation to create one from a downloaded Llama2 will be.

The Japanese-Korean task was a dataset created by merging multiple datasets and could be considered a novel artifact. However, as we do not have redistribution rights, we will publish only the sentence IDs for reproducibility purposes. 

\begin{table*}
\small
\begin{verbatim}
# Chinese (bleu, chrf, ter)
nrefs:1|case:mixed|eff:no|tok:zh|smooth:exp|version:2.4.2
nrefs:1|case:mixed|eff:yes|nc:6|nw:0|space:no|version:2.4.2
nrefs:1|case:lc|tok:tercom|norm:no|punct:no|asian:yes|version:2.4.2
\end{verbatim}

\begin{verbatim}
# Japanese (bleu, chrf, ter)
nrefs:1|case:mixed|eff:no|tok:ja-mecab-0.996-IPA|smooth:exp|version:2.4.2
nrefs:1|case:mixed|eff:yes|nc:6|nw:0|space:no|version:2.4.2
nrefs:1|case:lc|tok:tercom|norm:no|punct:no|asian:yes|version:2.4.2
\end{verbatim}

\begin{verbatim}
# Korean (bleu, chrf, ter)
nrefs:1|case:mixed|eff:no|tok:ko-mecab-0.996/ko-0.9.2-KO|smooth:exp|version:2.4.2
nrefs:1|case:mixed|eff:yes|nc:6|nw:0|space:no|version:2.4.2
nrefs:1|case:lc|tok:tercom|norm:no|punct:no|asian:yes|version:2.4.2
\end{verbatim}

\begin{verbatim}
# Bytes (bleu, chrf)
nrefs:1|case:mixed|eff:no|tok:none|smooth:exp|version:2.4.2
nrefs:1|case:mixed|eff:yes|nc:6|nw:0|space:no|version:2.4.2
\end{verbatim}

\caption{sacreBLEU signatures used for score computation.}
    
\end{table*}

\end{document}